\documentclass{article}
\usepackage[square,sort,comma,numbers]{natbib}

\usepackage[preprint]{nips_2018}

\usepackage[utf8]{inputenc} 
\usepackage[T1]{fontenc}    
\usepackage{hyperref}       
\usepackage{url}            
\usepackage{booktabs}       
\usepackage{amsfonts}       
\usepackage{nicefrac}       
\usepackage{microtype}      
\usepackage{algorithm}
\usepackage{algorithmic}
\usepackage[pdftex]{graphicx}
\usepackage{subfig}

\newtheorem{definition}{Definition}
\newcommand{\argmax}{\mathop{\rm arg~max}\limits}

\let\oldenumerate\enumerate
\renewcommand{\enumerate}{
   \oldenumerate
   \setlength{\itemsep}{1pt}
   \setlength{\parskip}{0pt}
   \setlength{\parsep}{0pt}
}
\let\olditemize\itemize
\renewcommand{\itemize}{
   \olditemize
   \setlength{\itemsep}{1pt}
   \setlength{\parskip}{0pt}
   \setlength{\parsep}{0pt}
}

\title{Lightweight Lipschitz Margin Training for \\ Certified Defense against Adversarial Examples}

\author{
  Hajime Ono\\
  University of Tsukuba\thanks{This work was done when the author was intern at NEC Corporation.}\\
  \texttt{hajime@mdl.cs.tsukuba.ac.jp} \\
  \And
  Tsubasa Takahashi\\
  NEC Corporation\\
  \texttt{t-takahashi@nk.jp.nec.com} \\
  \And
  Kazuya Kakizaki\\
  NEC Corporation\\
  \texttt{k-kakizaki@ay.jp.nec.com} \\
}

\begin{document}

\maketitle

\begin{abstract}
How can we make machine learning provably robust against adversarial examples in a scalable way?
Since certified defense methods, which ensure $\epsilon$-robust, consume huge resources, they can only achieve small degree of robustness in practice.
Lipschitz margin training (LMT) is a scalable certified defense, but it can also only achieve small robustness due to over-regularization.
How can we make certified defense more efficiently?
We present LC-LMT, a light weight Lipschitz margin training which solves the above problem.
Our method has the following properties;
(a) efficient: it can achieve $\epsilon$-robustness at early epoch, and
(b) robust: it has a potential to get higher robustness than LMT.
In the evaluation, we demonstrate the benefits of the proposed method.
LC-LMT can achieve required robustness more than 30 epoch earlier than LMT in MNIST,
and shows more than 90 $\%$ accuracy against both legitimate and adversarial inputs.
\end{abstract}

\section{Introduction}

Adversarial example is a crafted input for deceiving deep neural networks \cite{szegedy2014intriguing}.
It is a potentially critical safety issues in machine learning based autonomous systems like self-driving cars.
How can we make machine learning provably robust against adversarial examples?

One of simple defense approach against adversarial examples is to mask gradients \cite{papernot2016distillation}.
However, it provides a false sense of security \cite{carlini2017towards} \cite{athalye2018obfuscated} \cite{tramer2018ensemble}.
Adversarial training \cite{goodfellow2015explaining} \cite{kurakin2017adversarial} \cite{madry2018towards}, which injects adversarial examples with correct labels into training samples, is a promising approach, but it is just a countermeasure against collectable samples.

Recently, several certified robust learning approaches have been proposed \cite{wong2018provable} \cite{mirman2018differentiable}, but all of them lack scalability.
Thus, in a practical time, those method can only achieve small degree of robustness even for low dimensional datasets.
Lipschitz margin training (LMT) \cite{tsuzuku2018lipschitz} is a scalable certified defense, but it can also only achieve small degree of robustness due to over-regularization.
How can we make certified defense more efficiently?

We present LC-LMT, a lightweight Lipschitz margin training which solves the above problem.
The contributions of the proposed method are:
(a) efficient: it can achieve $\epsilon$-robustness at early epoch, and
(b) robust: it has a potential to get higher robustness than LMT.
In the evaluation, we demonstrate the benefits of the proposed method.
LC-LMT can achieve required robustness more than 30 epoch earlier than LMT in MNIST,
and shows more than 90 $\%$ accuracy against both legitimate and adversarial inputs.

\section{Preliminary}

Here, we introduce essential notations, definitions and existing work to understand our proposal.

\subsection{$\epsilon$-robustness}
$\epsilon$-robustness is a certifiable metric representing robustness of a neural network against adversarial examples.
\begin{definition}{($\epsilon$-robustness)}
  Let $B^{p}_{\epsilon}(x)=\{x+\delta|\|\delta\|_p \leq \epsilon\}$ denote the $\ell_p$-ball of radius $\epsilon$ around a point $x \in \mathbb{R}^d$. A neural network $f$ is called $\epsilon$-robust around a point $x$ if $f$ assigns the same class to all points $\tilde{x} \in B^{p}_{\epsilon}(x)$.
\end{definition}
This paper considers $\epsilon$-robustness in $\ell_2$-norm ($p=2$) as well as \cite{tsuzuku2018lipschitz}.

\subsection{Lipschitz Margin Training}

Lipschitz Margin Training (LMT) is a scalable way to ensure $\epsilon$-robustness based on the Lipschitz constant \cite{tsuzuku2018lipschitz}.
Note that, we assume the last layer of $f$ is softmax, and $f(x)$ represents logits which is output of the subnetwork before the softmax.

For all $x$ and $x+\delta$, a neural network $f$ is Lipschitz continuous if there exists a real constant $L_f \geq 0$ such that
\begin{equation}
 \|f(x)-f(x+\delta)\|_2 \leq L_f \|\delta\|_2.
\end{equation}
Lipschitz constant $L_f$ represents sensitivity of $f$ if the input is changed at most $\delta$.

LMT also introduced a remarkable notation of prediction margin.
\begin{definition}{(Prediction Margin)}
  Let $x$ be an input image and $t$ be the true label of $x$, the prediction margin of $x$ in $f$ is computed as:
  \begin{equation}
    M_{f,x}:=f(x)_t - max_{i \neq t} f(x)_i .
    \label{eq:margin}
  \end{equation}
\end{definition}

LMT enlarges the prediction margin around each input $x$ to be $\epsilon$-robustness.
The prediction margin to satisfy $\epsilon$-robustness for $x$ is derived in \cite{tsuzuku2018lipschitz} as follows:
\begin{equation}
  M_{f,x} \geq \sqrt{2} L_f \epsilon .
  \label{eq:ineq-margin}
\end{equation}

Therefore, we need at least $\beta = \sqrt{2} L_f \epsilon$ between true class and the others to make $f$ around each $x$ $\epsilon$-robust.
In the training phase, LMT inflates each $f(x)_i$ except $f(x)_t$.
The algorithm of LMT is described in Algorithm \ref{alg:lmt}.

The above inflation has an effect like regularization towards obtaining the margin to be $\epsilon$-robust.
LMT can achieve $\epsilon$-robustness in a scalable way through this regularized training.
However, we found that it has an issue like over-regularization in case either $\epsilon$ and $L_f$ is large.
Due to the issue, LMT tends to slowly forward training.
Thus, it consumes lots of epochs to get the required margin and may cause under-fitting.
Actually, LMT is hard to achieve $\epsilon$-robustness with $\epsilon \geq 0.5$ even in MNIST.

\section{Proposed Method}
We propose low-cost Lipschitz margin training (LC-LMT) to solve the above problem.
We introduce a limited inflation to make network $f$ $\epsilon$-robust more efficiently.

Let $v$ be the class whose score $f(x)_v$ is the highest in all $f(x)_i$ where $i \neq t$.
If the gap $f(x)_t - f(x)_v$ is larger than $\beta$, all gaps between $t$ and $i(\neq t)$ satisfy (\ref{eq:ineq-margin}).

\begin{minipage}[t]{0.48\hsize}
  \begin{algorithm}[H]
    \small
    \caption{Lipschitz Margin Training \cite{tsuzuku2018lipschitz}}
    \label{alg:lmt}
    \begin{algorithmic}[1]
      \REQUIRE $\epsilon$, image $x$, label $t$
      \STATE $y \leftarrow$ Forward$(x)$
      \STATE $L_f \leftarrow$ CalcLipschitzConst()
      \FORALL{$i \neq t$}
        \STATE $y_i \leftarrow y_i + \sqrt{2}L_f \epsilon$
      \ENDFOR
      \STATE $p \leftarrow$ SoftmaxIfNeccesary$(y)$
      \STATE $\ell \leftarrow$ CalcLoss$(p, t)$
    \end{algorithmic}
  \end{algorithm}
\end{minipage}
\hfill
\begin{minipage}[t]{0.48\hsize}
  \begin{algorithm}[H]
    \small
    \caption{LC-LMT}
    \label{alg:lclmt}
    \begin{algorithmic}[1]
      \REQUIRE $\epsilon$, image $x$, label $t$
      \STATE $y \leftarrow$ Forward$(x)$
      \STATE $L_f \leftarrow$ CalcLipschitzConst()
      \STATE $v \leftarrow \argmax_{i \neq t} f(x)_i$
      \STATE $y_v \leftarrow y_v + \sqrt{2}L_f \epsilon$
      \STATE $p \leftarrow$ SoftmaxIfNeccesary$(y)$
      \STATE $\ell \leftarrow$ CalcLoss$(p, t)$
    \end{algorithmic}
  \end{algorithm}
\end{minipage}

Based on the above idea, we freight the inflation value $\beta$ only at $v$.
In the training phase, the inflation has an effect of increasing the score $f(x)_t$ to be $\epsilon$-robust around $x$ (Algorithm \ref{alg:lclmt}).

Our proposed method LC-LMT only inflates the score $f(x)_v$, but LMT inflates all except $t$.
Then, the rank of $t$ in $f(x)$ can be changed at most 1 in LC-LMT, at most number of classes - 1 in LMT.
Since softmax activation tends to depress lower rank values into zero, the chance of the depression of $t$ in LMT is higher than LC-LMT.
Thus, in terms of regularization, the regularization cost of LC-LMT is thought of as smaller than LMT.
The volume of the cost may be related with sum of inflated values and
number of inflated classes.
The theoretical analysis about the above discussion is a future work.

\section{Evaluation}
In this section we demonstrate the effectiveness of the proposed method LC-LMT.
The experiments were designed to answer following questions:
\begin{itemize}
\item[Q1] \textit{Efficiency}: How early is our method in achieving $\epsilon$-robustness?
\item[Q2] \textit{Robustness}: How accurate is our method in classifying adversarial inputs?
\end{itemize}

\noindent
\textbf{Experimental setting.}
We used MNIST and SVHN \cite{netzer2011reading} for the evaluations.
For MNIST, we used a neural network having 4 fully-connected layer with ReLU activation and softmax at output layer.
Each hidden layer has 100 parameters.
For SVHN, we used WideResnet \cite{zagoruyko2016wide} with 16 layers and width factor 4 following \cite{cisse2017parseval}.
Computation of Lipschitz constant follows LMT.
To generate adversarial examples, we employed (L2-) Carlini Wagner attack (CW attack) \cite{carlini2017towards} with 100 iterations.

\subsection{Efficiency}

\begin{figure}[b]
  \centering
  \subfloat[Margin@epoch]{
    \includegraphics[width=0.45\hsize]{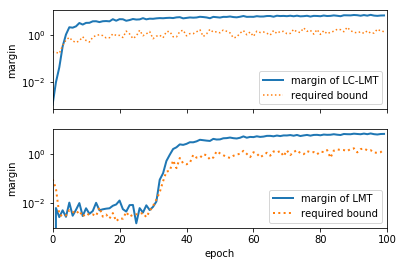}
    \label{fig_margin}
  }\hfill
  \subfloat[Loss@epoch]{
    \includegraphics[width=0.45\hsize]{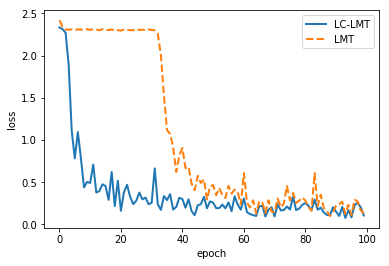}
    \label{fig_loss}
  }
  \caption{LC-LMT is efficient. LC-LMT achieves (\ref{eq:ineq-margin}) at early epoch.}
\end{figure}

We now evaluate efficiency of proposed method. Here, we employed MNIST and $\epsilon=1.0$.

First, we show how early is our method in enlarging margin which satisfies (\ref{eq:ineq-margin}).
Figure \ref{fig_margin} shows the enlarged margin and required bound in average for each method.
LC-LMT has been satisfied the required bound since 5th epoch.
While, LMT satisfied it around 35th epoch.
Thus, our proposed method LC-LMT can satisfy (\ref{eq:ineq-margin}) at very early epoch.

Next, we show how early is our method in reducing loss.
Figure \ref{fig_loss} shows loss value at each epoch.
LC-LMT shows dramatical loss drops at very early epoch similar to non-robust learning.
However, LMT cannot reduce loss in first 30 epoch.
Since the difference between both methods are number of inflations on logits, the size of overall inflations might be a key to be efficient.

The above two results suggest that our proposed method provide efficient robust training.

\begin{figure}[t]
  \centering
  \subfloat[MNIST]{
    \includegraphics[width=0.45\hsize]{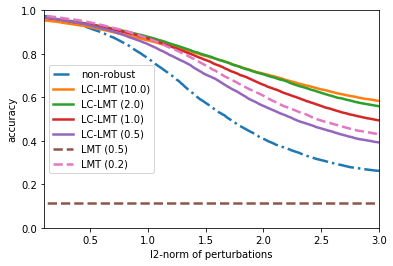}
    \label{fig_acc_mnist}
  }\hfill
  \subfloat[SVHN]{
    \includegraphics[width=0.45\hsize]{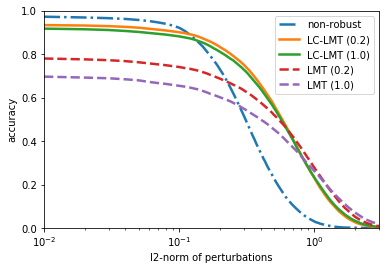}
    \label{fig_acc_svhn}
  }
  \caption{LC-LMT is robust. LC-LMT classifies adversarial inputs more correctly.}
\end{figure}

\subsection{Robustness}

\begin{table}[t]
  \small
  \centering
  \caption{Accuracy against legitimate inputs [SVHN]}
  \label{tbl_acc}
  \begin{tabular}{l|c|c|c|c|c}
              & LMT (0.2)  & LMT (1.0) &  LC-LMT (0.2)  & LC-LMT (1.0)  & non-robust \\ \hline
    Accuracy (Train) & 0.500        &  0.335      & 0.933            & 0.911 & 0.984 \\
    Accuracy (Test) & 0.785        &  0.700      & 0.936            & 0.912 & 0.975 \\
  \end{tabular}
\end{table}

Next, we measure robustness.
Figures \ref{fig_acc_mnist} and \ref{fig_acc_svhn} plot classification accuracy against adversarial examples along with $\ell_2$-norm of perturbations.
We run LMT and LC-LMT in 100 epochs.

Figure \ref{fig_acc_mnist} shows accuracy against adversarial examples for MNIST.
LC-LMT shows high accuracy as well as non-robust learning against very small perturbed inputs whose norm of perturbations is less than 1.
LMT (0.2), which is LMT at $\epsilon=0.2$, also shows good accuracy, but LMT (0.5) cannot correctly classify at all.
For large perturbed inputs, LC-LMT shows good accuracy in proportional to $\epsilon$.
However, LC-LMT (2.0) and LC-LMT (10.0) do not satisfy their robustness demands.
Those demands might be too large to build classifiers with using MNIST.
Anyway, those results suggest that LC-LMT has a potential to make machine learning more robust than LMT.
LC-LMT still works at $\epsilon=10$ for MNIST.

On SVHN, Figure \ref{fig_acc_svhn} shows that LC-LMT get higher accuracy than LMT until norm of perturbations < 0.7, but LMT is slightly better  after 0.7.
Table \ref{tbl_acc} shows training accuracy and test accuracy in each setting.
The table suggests that LC-LMT shows accuracies close to both training data and test data, but LMT is neither of them.
The above results demonstrate that our proposed method LC-LMT is accurate for both legitimate and adversarial inputs.

\section{Conclusion}
We proposed LC-LMT, a low cost Lipschitz margin training.
Our method has the following advantages;
(a) efficient: it can achieve $\epsilon$-robustness at early epoch, and
(b) robust: it has a potential to get higher robustness than LMT.
Evaluation showed that LC-LMT achieved required robustness in very early epochs, and it demonstrated more than 90 $\%$ accuracy against both legitimate and adversarial inputs.
In the future works, we tackle theoretical analysis of the proposed method and the relationships between LMT and the other Lipschitz based works.

\clearpage

\small
\bibliographystyle{abbrv}

\end{document}